\documentclass[runningheads]{llncs}
\usepackage{graphicx}
\usepackage{color}
\usepackage{amsmath}
\usepackage{multirow}
\usepackage{multicol}
\usepackage{mathtools}
\usepackage{mathrsfs}
\usepackage{caption}

\usepackage{subfig}
\usepackage{amsmath,amssymb,amsfonts}
\usepackage{algorithmic}
\usepackage{graphicx}
\usepackage{textcomp}
\usepackage{algorithm}
\usepackage{graphics}
\usepackage{threeparttable}
\usepackage{color}
\usepackage[normalem]{ulem}
\usepackage{multirow}
\usepackage{float}
\usepackage{amsfonts}
\usepackage{bm}
\usepackage{array}
\usepackage{colortbl}
\usepackage{pifont}
\usepackage{diagbox}
\usepackage{rotating}
\usepackage{booktabs}
\usepackage{overpic}
\usepackage{makecell}
\usepackage{contour}
\usepackage{adjustbox}
\usepackage[misc]{ifsym} 

\def\ourmodel{TriTemp-OR}
\begin{document}

\newcommand{\tabref}[1]{Table \ref{#1}}
\newcommand{\tickYes}{\bullet}
\newcommand{\cmark}{\ding{51}}
\newcommand{\tickNo}{\hspace{1pt}\ding{55}}
\newcommand{\figref}[1]{Fig. \ref{#1}}
\newcommand{\supp}[1]{\textcolor{magenta}{#1}}
\newcommand{\notsure}[1]{{\textcolor{red}{#1}}}
\newcommand{\secref}[1]{Sec. \ref{#1}}
\newcommand{\sArt}{state-of-the-art}
\newcommand{\jl}{{\color{red} \bf Jialun}}
\newcommand{\dd}{{\color{green} \bf Diandian}}
\newcommand{\mx}[1]{{\color{cyan} \bf Manxi: #1}}  

\newcommand{\ie}{{\emph{i.e.}}}
\newcommand{\viz}{{\emph{viz.}},\xspace}
\newcommand{\eg}{{\emph{e.g.}}}
\newcommand{\etc}{etc.}
\newcommand{\etal}{{\emph{et al.}}}

\title{Tri-modal Confluence with Temporal Dynamics for Scene Graph Generation in Operating Rooms}
\titlerunning{TriTemp-OR}
\author{Diandian Guo\thanks{Equal contribution.}$^{1}$, Manxi Lin\inst{\star}$^{2}$, Jialun Pei\,$^{\textrm{\Letter}1}$, He Tang$^{3}$, Yueming Jin$^{4}$, \ \ \ \ \ 
Pheng-Ann Heng$^{1}$}
\institute{The Chinese University of Hong Kong$^{1}$, Technical University of Denmark$^{2}$, Huazhong University of Science and Technology$^{3}$, National University of Singapore$^{4}$}
\authorrunning{Diandian Guo et al.}
%参考MICCAI Submission Q&A，不应移除作者和单位，而应使用星号代替

\maketitle         
\begin{abstract}

A comprehensive understanding of surgical scenes allows for monitoring of the surgical process, reducing the occurrence of accidents and enhancing efficiency for medical professionals. 
Semantic modeling within operating rooms, as a scene graph generation (SGG) task, is challenging since it involves consecutive recognition of subtle surgical actions over prolonged periods. 
To address this challenge, we propose a \textbf{Tri}-modal (\ie, images, point clouds, and language) confluence with \textbf{Temp}oral dynamics framework, termed \ourmodel. 
Diverging from previous approaches that integrated temporal information via memory graphs, our method embraces two advantages: 1) we directly exploit bi-modal temporal information from the video streaming for hierarchical feature interaction, and 2) the prior knowledge from Large Language Models (LLMs) is embedded to alleviate the class-imbalance problem in the operating theatre.
Specifically, our model performs temporal interactions across 2D frames and 3D point clouds, including a scale-adaptive multi-view temporal interaction (ViewTemp) and a geometric-temporal point aggregation (PointTemp).
Furthermore, we transfer knowledge from the biomedical LLM, LLaVA-Med, to deepen the comprehension of intraoperative relations. 
The proposed \ourmodel~enables the aggregation of tri-modal features through relation-aware unification to predict relations so as to generate scene graphs.
Experimental results on the 4D-OR benchmark demonstrate the superior performance of our model for long-term OR streaming. The code will be publicly available.
\end{abstract}

\section{Introduction}
With the growing complexity and diversity of modern surgical procedures, the automated scene understanding of the operating room (OR) using computer-assisted systems has become a critical necessity~\cite{4dor,4dor_article}.
Unlike specific surgical assistance interventions, \eg, surgical phase recognition \cite{gao2021trans}, instrument segmentation~\cite{sestini2023fun}, and anatomy tracking~\cite{green2018first}, holistic scene modeling of operating theatres \cite{4dor,LABRADOR,4dor_article} can facilitate coordination and communication among surgical teams, optimize the surgical process, and enhance safety and efficiency during surgery.
Scene graph generation (SGG)~\cite{chang2021comprehensive,wald2022learning,cong2023reltr}, by condensing image content through nodes and edges along with their relationships, enables effective monitoring of surgical procedure and detailed guidance for operating activities~\cite{4dor}. 
However, generating scene graphs within the complex OR environment is a challenging task that requires relational associations between subjects and objects (\eg, surgeons, patients, and medical equipment) in ORs and involves continuous follow-up of operation details throughout the surgical period.

To address this challenge, the pioneering work 4D-OR~\cite{4dor} adopts a multi-stage framework, initially identifying medical staff and scene objects via a 3D pose estimation model and an object detector, followed by a modified 3DSGG network~\cite{wald2020learning} to generate scene graphs.
Nevertheless, this OR scene graph generation (OR-SGG) approach neglects to explore temporal information to enhance graph construction.
LABRAD-OR~\cite{LABRADOR} introduces a memory scene graph to represent dynamic temporal characteristics and combines visual information extracted from 2D and 3D modalities to achieve more consistent predictions.
However, the multi-stage spatial-temporal fusion, \ie, first generating a coarse graph and then conducting sequential interaction, diminishes the sensitivity to fine-grained visual features and exploitation of spatial-temporal information. For better utilizing temporal dynamics, we suggest directly leveraging bi-modal temporal cues for finer-grained hierarchical feature interaction. 
Since the process of surgery exhibits wide variations in the duration of activities at different phases, it is inevitable to confront a significant imbalance of classes between frequent ones (\eg, `Lying on') and rare ones (\eg, `Cutting'). 
% We identify another essential property of surgical procedure that different activities exhibit high variability in their temporal duration. This is inevitable to confront a significant class imbalance issue, \eg, `Lying on' (frequent) and `Cutting' (rare) in knee surgery.
% Recent works in similar tasks~\cite{ovsgg,vlsat,ding2023pla} reveal the knowledge from large language models (LLMs)~\cite{llava,flamingo,singhal2023large} mitigates the class-imbalance problem. The enriched and subtle semantic representations that LLMs learn from large-scale pre-training rebalance the importance of rare categories. Inspired by these methods and considering the knowledge gap between open vocabulary and medical semantics, we reformat each triplet $<$subject, predicate, object$>$ in the graph into a proper template and then leverage prior knowledge in LLaVA-Med~\cite{llavamed}, an LLM fine-tuned on biomedical data,  to mitigate the class imbalance problem via the feature alignment strategy.
Several natural scene graph models~\cite{ovsgg,vlsat,ding2023pla} employ knowledge distillation from large language models (LLMs)~\cite{llava,flamingo,singhal2023large} pre-trained on large amounts of textual data to provide enriched and subtle semantic representations to improve the prediction accuracy for rare classes.
While LLMs have shown considerable effectiveness in various semantic scene modeling tasks~\cite{genvlkt,gao2023compositional,liu2023clip} depending on the broad understanding of language and context, the knowledge gap between open vocabulary and medical semantics hinders the application of LLMs in the OR-SGG domain. 
% To bridge this gap, we propose a knowledge transfer scheme, leveraging LLaVA-Med~\cite{llavamed}, an LLM fine-tuned on biomedical data, to mitigate the class-imbalance problem via feature alignment strategy.
In this regard, it is appropriate to utilize LLMs fine-tuned on biomedical data to mitigate the class-imbalance problem.

In this work, we propose an end-to-end tri-modal OR-SGG model, named \ourmodel.
Our model first performs hierarchical feature interaction with 2D and 3D temporal information in video streaming through a scale-adaptive multi-view temporal interaction (ViewTemp) and a geometric-temporal point aggregation (PointTemp).
ViewTemp embraces a scale-adaptive feature partition strategy based on the receptive fields from different views, to encourage multi-scale feature identification in multi-view 2D video streaming.
In parallel, PointTemp captures 3D point cloud temporal information through point 4D convolutions and global temporal aggregation with self-attention~\cite{p4former,attentionisallyouneed}.
Concentrating on subject-object interaction spaces, we also introduce a relation-aware feature unification operation to aggregate 2D spatial-temporal and 3D geometric-temporal information.
% subject-object interaction areas.
More importantly, the proposed~\ourmodel~aligns multi-view image and point cloud features with textual semantic features distilled from the biomedical pre-trained LLM (\ie, LLaVA-Med~\cite{llavamed}) during training, yielding effective relation-aware tri-modal representations for predicting subject-object relations and generating scene graphs.
Experimental results demonstrate that \ourmodel~achieves superior performance on the 4D-OR dataset, indicating the potential application value in surgical intervention and assistance.

\section{Method}

\begin{figure*}[t!]
	\centering
	\includegraphics[width=\linewidth]{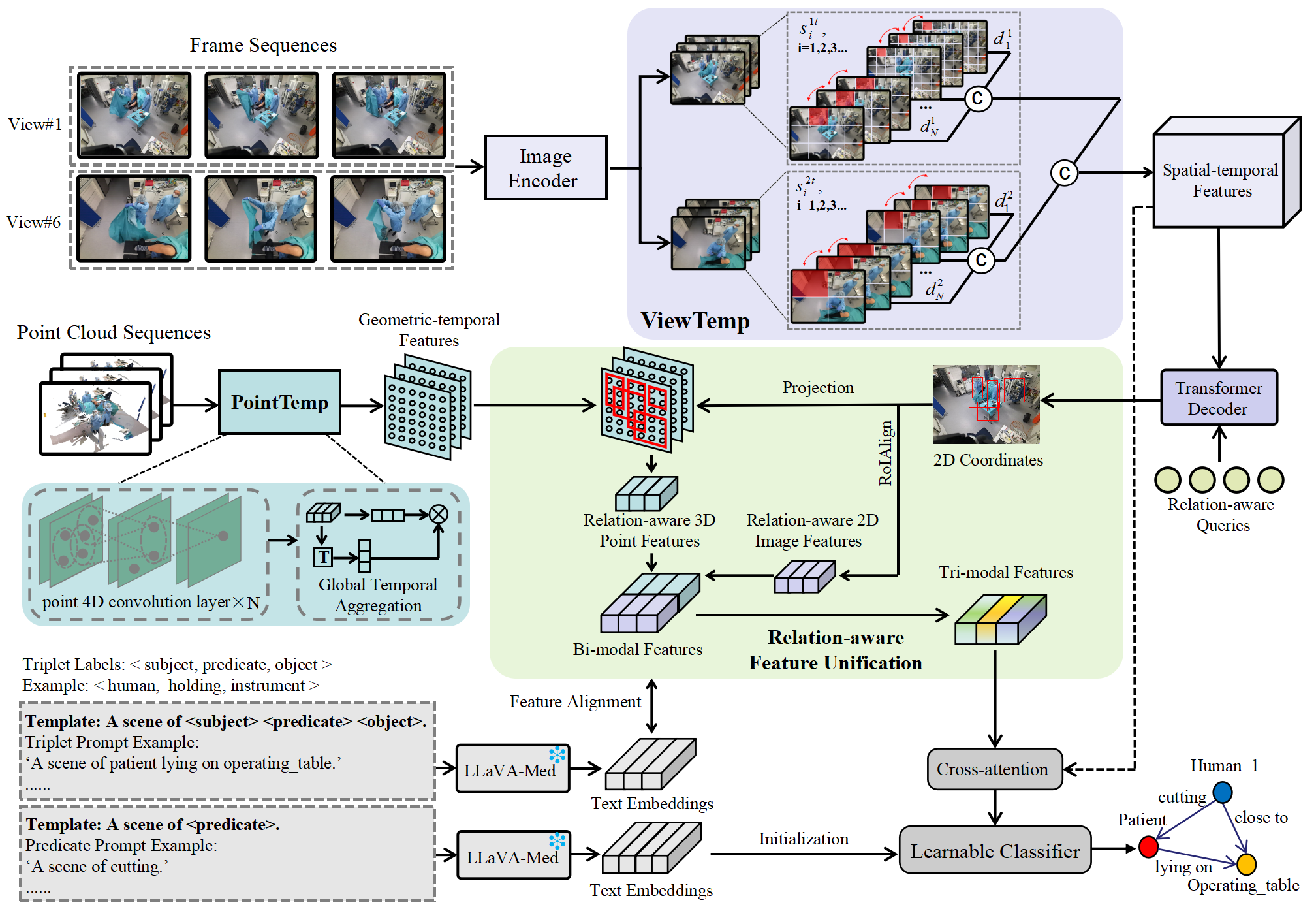}
        % \vspace{-8pt}
	\caption{Overview of the proposed \ourmodel~for scene graph generation in ORs.}
	\label{main}
    % \vspace{-10pt}
\end{figure*}
% Overview of the proposed \ourmodel~for scene graph generation in operating rooms, with the illustration of ViewTemp, PointTemp, and knowledge transfer from LLaVA-Med \cite{llavamed}.

% We extract temporal 2D features from a CNN encoder, and integrate them through view-aware scale-adaptive temporal fusion (ViewTemp) for temporal features, which are fed into the transformer decoder for potential relation features and corresponding 2D coordinates. From 2D coordinates, we generate relation-aware 2D and 3D triplet features.

% \vspace{-5pt}
\figref{main} illustrates our tri-modal confluence with temporal dynamics framework \ourmodel~for generating scene graphs in operating rooms.
We first utilize a ResNet-50~\cite{resnet} to extract spatial features from multi-view video frames.
Then, we leverage the proposed scale-adaptive multi-view temporal interaction (ViewTemp) and geometric-temporal point aggregation (PointTemp) to facilitate temporal interactions for 2D multi-view features and 3D point clouds.
The spatial-temporal features derived from ViewTemp are then fed into the transformer decoder with relation-aware queries to predict 2D subject-object coordinates.
% This scheme provides efficient integration of 2D and 3D features in relation-aware locations for bi-modal feature representation.
The workflow provides efficient bi-modal feature representations in relation-aware locations.
Furthermore, we reformat triplets $<$subject, predicate, object$>$ from the graph into a proper template for input to the frozen LLaVA-Med~\cite{llavamed} to distill text embeddings, which can enhance the understanding of the intraoperative activities by alignment with bi-modal features.
After that, the relation-aware feature unification is introduced to aggregate 2D and 3D features into subject-object representations for paired relationship prediction.
Finally, we interact relation-aware tri-modal representations (\ie, images, point clouds, and language) with spatial-temporal features via cross-attention, embracing a learnable classifier initialized by predicate text embeddings from LLaVA-Med, to generate scene graphs.

\subsection{Multi-view Image and Point Cloud Temporal Dynamics}

% The temporal information helps capture the dynamics between consecutive frames that depict static scenes. Surgical environments, which are typically characterized by various entities including doctors and nurses, as well as surgical behaviors like cutting and cementing \mx{we need to make sure cutting and cementing are "surgical terms"}, pose requirements on the accurate identification and tracking of the dynamics. 

% For viewpoints with coarse-grained features, we apply coarser granularity in feature partitions and coarser local attention strategy, ensuring consistent feature correspondence across frames.
\subsubsection{ViewTemp.}
% Utilizing multi-view temporal information enables leveraging data captured from multiple perspectives, offering a comprehensive spatial-temporal representation in OR scenes. The integration of temporal information from multiple viewpoints contributes to better depth perception and 3D modeling of surgical environments across different frames and angles. Besides, multiple cameras are set from various viewpoints for capturing complex surgical scenes in operating rooms, where different views exhibit diverse visual granularities according to their camera poses. Take the 4D-OR~\cite{4dor} dataset for example, view\#1 has a broader horizon of vision, while view\#6 provides a more focused and finer-grained display of the surgery. 
% Therefore, we introduce the scale-adaptive multi-view temporal interaction (ViewTemp) for the temporal integration of 2D image features across various viewpoints. We combine the dense and sparse partition to mine finer features for the fine-grained viewpoints, while a sparser patch partition approach is adopted in other viewpoints to adaptively recognize patterns at different feature scales. ViewTemp runs varying-scale convolutional kernels tailored to each viewpoint's granularity, followed by local temporal attentions across neighboring frames for temporal aggregation. 
Temporal dynamics of 2D frames can facilitate the consistency of spatial features at consecutive time points.
% To capture complex scenes in the OR, the 4D-OR dataset contains videos from different viewpoints, presenting varying receptive fields and visual granularities.
% For instance, view\#1 offers a broader field of view, while view\#6 allows for more focused and finer-grained observation of surgery.
To this end, we introduce the scale-adaptive multi-view temporal interaction (ViewTemp) for spatial-temporal integration of 2D image features across various viewpoints.
Given six viewpoints in the 4D-OR dataset, we select view\#1 to offer a broader field of view and view\#6 to allow for finer-grained observation of the surgery.
ViewTemp operates varying-scale convolution kernels depending on the granularity of each viewpoint, which explores finer features in view\#1 and interacts with macro-level cues in view\#6.
% This operation allows for adaptive temporal aggregation of consecutive frames within different scales of feature space.
%
Specifically, we predefine a view-specific set of $N$ various-size kernels as $\mathbb{M}_k \coloneqq \left \{m_{i}^{k}\right \}_{i=1}^{N}$ for the $k$-th viewpoint, where $m_{i}^{k}$ is determined according to the feature granularity. 
Given the feature sequence from ResNet-50~\cite{resnet} encoder for the $k$-th viewpoint, we extract a group of multi-scale features $\mathbb{S}^{k}_{t}\coloneqq\left \{ s_{i}^{kt}\right \}_{i=1}^{N}$ for each feature map $f_{t}^{k}$ at time point $t$ by applying various-size kernels:
\begin{equation}\footnotesize\label{Eq:1}
s_{i}^{kt} =  f_{t}^{k} \ast m_{i}^{k},
\end{equation}
where $i$ represents the $i$-th convolution kernel in $\mathbb{M}_{k}$ and $*$ denotes the convolution operation. 
Afterward, for each feature scale, the image features $s_{i}^{kt}$ at different time points $t$ are interacted through the local cross-attention \cite{attentionisallyouneed} for hierarchical feature extraction. We denote a set of the same-scale features from consecutive video frames as $\mathbb{U}_{i}^{kt} \coloneqq \left \{s_{i}^{kj}\right \}_{j=t-l+1}^{j=t}$, where $l$ is the number of consecutive frames in one batch.
% We use $s_{i}^{kt}$ as queries and $\mathbb{U}_{i}^{kt}$ as keys and values for local interaction following:
Here, $s_{i}^{kt}$ is used as queries and $\mathbb{U}_{i}^{kt}$ serves as keys and values for local interaction:
\begin{equation}\footnotesize\label{Eq:gene_qkv}
    Q=\mathcal{W}_q(s_{i}^{kt}),\quad K=\mathcal{W}_k({\mathcal{C}}(\mathbb{U}_{i}^{kt})), \quad V=\mathcal{W}_v({\mathcal{C}}(\mathbb{U}_{i}^{kt})),
\end{equation}
\begin{equation}\footnotesize\label{Eq:dynamic}
    {d}_{i}^{k} = \emph{CA}(Q, K, V),
\end{equation}
where $\mathcal{W}(\cdot)$ represents the fully connected layer and ${\mathcal{C}}(\cdot)$ is the concatenation operation. \emph{CA}$(\cdot)$ means the local cross-attention operation, where only feature points with consistent locations are interacted for temporal integration. ${d}_{i}^{k}$ denotes the temporal-fused features of the $i$-th scale. Then, we merge the features of different scales in each viewpoint by MLP to attain the multi-scale temporal representation.
% \begin{equation}\footnotesize\label{Eq:2}
%     {\Tilde{f}}_{k} = \text{MLP}(\mathcal{C}(\left \{{d}_{i}^{k}\right \}_{i=1}^{N})).
% \end{equation}
% The described multi-view temporal interaction self-adaptively integrates temporal features with diverse granularities across various viewpoints. The hierarchical feature interaction encourages multi-scale intraoperative action recognition for multi-view 2D video streaming with different camera poses in operating rooms.
The proposed ViewTemp adaptively aggregates multi-level granularity spatial-temporal features across various viewpoints.

% method not only efficiently extracts principal features across viewpoints but also ensures the alignment of multi-level features in consecutive frames, thereby enhancing the robustness of temporal information interaction.
\subsubsection{PointTemp.}
% Point cloud videos offer a detailed perspective on the dynamic interactions within the 3D world, capturing how objects move and interact with their surroundings. They are particularly valuable for recognizing surgical actions and for detailing global geometric dynamics in OR environments, where each viewpoint only exhibits limited visibility. Therefore, we design the geometric-temporal point aggregation (PointTemp) to explore the temporal dynamics in point cloud videos for the semantic modeling within operating rooms. Specifically, our PointTemp consists of point 4D convolution layers~\cite{p4former} and an additional global temporal aggregation. While the 4D convolution embeds spatial-temporal local structures presented in the point cloud video, the global temporal aggregation performs temporal interactions for point cloud features from various timestamps via self-attention \cite{attentionisallyouneed} operations to capture spatial-temporal correlations.
The point cloud provides a detailed 3D perspective of dynamic interactions, which is essential for recognizing subject-object relations and describing global geometric dynamics in ORs.
Thus, we design the geometric-temporal point aggregation (PointTemp) to explore temporal dynamics in 3D point clouds for more consistent alignment with 2D multi-view spatial-temporal features.
As illustrated in~\figref{main}, PointTemp consists of point 4D convolution layers and a global temporal aggregation~\cite{p4former}.
The 4D convolution layer perceives the local geometric-temporal structures within the point cloud, while the global temporal aggregation layer performs temporal interactions from various time points via self-attention~\cite{attentionisallyouneed} to capture geometric-temporal correlations.

\subsection{Relation-aware Feature Unification}

% Our OR-SGG model transcends traditional approaches by not only incorporating language embeddings but also by deploying an advanced heuristic-level multimodal, Relation-aware feature fusion mechanism. Following the application of View-aware Scale-adaptive Temporal Fusion, we derive temporal features from multi-view imagery. The subsequent phase involves a critical task: the integration of these temporal domain features with 2D images, 3D point clouds, and textual data for a comprehensive multimodal representation.

2D image and 3D point cloud feature representations are crucial for semantic modeling in ORs~\cite{4dor}.
To facilitate feature unification, we employ 2D spatial coordinates to integrate relation-aware bi-modal features.
In detail, spatial-temporal features from ViewTemp are passed through a transformer decoder to predict the locations of all entities.
Then, we acquire 2D relation-aware features in the target locations by RoIAlign~\cite{he2017mask}.
Meanwhile, we project the 2D coordinates into the 3D point cloud domain by applying the camera projection matrix for spatial translation.
For each subject-object pair, we extract the relation-aware point cloud features from geometric-temporal features through a max-pooling operation among neighborhood points. Finally, the relation-aware bi-modal features are aggregated into subject-object representations for paired relationship prediction. The proposed relation-aware feature unification utilizes 2D images and 3D point clouds accompanied by temporal dynamics for efficient bi-modal feature integration to enrich the representation of subject-object relationships.

\subsection{Knowledge Transfer from LLaVA-Med}

Since different phases in surgery pose varying durations, we consider exploiting the rich semantic representations and the broad understanding of language and context from LLMs to alleviate the class imbalance in ORs, \ie, frequent classes versus rare classes.
However, the knowledge gap between the open vocabulary and medical scenarios poses challenges for the application of LLMs in the OR-SGG task.
We alleviate this issue with the novel biomedical LLM, LLaVA-Med~\cite{llavamed}, in two steps, including knowledge distillation for each triplet in the graph and knowledge transfer for each predicate.
As shown in~\figref{main}, we first regularize the relation-aware bi-modal features using the corresponding text embeddings from LLaVA-Med.
Before training, we prompt all triplets in the graph with the template `A scene of a/an [subject][predicate] a/an [object]' and store the average of the frozen LLaVA-Med token features from the last layer.
Then, the bi-modal features are aligned with the transferred token features using L1 loss.
After aggregating the bi-modal features with textual features via relation-aware feature unification, the unified tri-modal features are passed through a cross-attention operation followed by a learnable classifier.
Notably, the classifier is initialized with text embeddings generated by LLaVA-Med from the template `A scene of [predicate]' to predict relationships.
Considering the fixed triplet combinations, we extract token features from the pre-trained LLaVA-Med offline before training.
Additionally, the alignment between bi-modal features and text embeddings is performed during training. 
Overall, we efficiently transfer prior knowledge from medical LLMs into our model by imposing additional constraints on relation-aware features, alleviating the class imbalance in ORs.

Accordingly, the total loss function of our model is expressed as
\begin{equation}\footnotesize
\mathcal{L}_{\text{total}}=\mathcal{L}_{coord}+\lambda_{c}\mathcal{L}_{cls}+\lambda_{t}\mathcal{L}_{text},
\end{equation}
where $\lambda_{c}$ and $\lambda_{t}$ are weighting factors. The coordinate loss $\mathcal{L}_{coord}$ includes L1 loss and GIoU loss \cite{giou} for coordinate prediction and focal loss \cite{focal} for predicted subject/object scores. The relation classification loss $\mathcal{L}_{cls}$ can be computed as
\begin{equation}\footnotesize
L_{cls} = \frac{1}{\sum_{i=1}^N  \sum_{z=1}^Z y_{iz}} \sum_{i=1}^N \sum_{z=1}^Z F(\hat{y}_{iz}, y_{iz}),
\end{equation}
where $F(\cdot)$ denotes focal loss and $Z$ is the number of relation classes. $N$ denotes the number of predicted subject-object pairs. $y_{iz} \in \{0,1\}$ indicates whether the $i$-th prediction contains the $z$-th relation class. 
$\hat{y}_{iz}$ is the predicted score of the $z$-th relation class. $\mathcal{L}_{text}$ is the L1 loss between bi-modal features and text embeddings from LLaVA-Med for alignment.
% \begin{equation}\footnotesize
% \mathcal{L}_{text} = \frac{1}{n}\sum\left |f_{bm}-f_{llava}  \right |.
% \end{equation}
% Given the limited number of possible relation triplets in our work, the features from the pre-trained LLaVA-Med is offline extracted without burdening the model with extensive parameters. Additionally, the alignment between the relation-aware bi-modal features and text embeddings is restricted to the training phase. We directly utilize the aligned features for prediction during inference, further reducing the computational load. In summary, by imposing additional constraints on the relation-aware features, we transfer knowledge from medical LLMs to alleviate the class imbalance problem in OR scenes.

% \vspace{-5pt}
\section{Experiments}
\subsection{Dataset and Evaluation Metrics}
% We validated the proposed framework on the large-scale public benchmark 4D-OR~\cite{4dor}. It consists of 10 videos of knee replacement surgery procedures, involving 6,734 frames captured by RGB-D cameras. We consider each frame a scene, described by six 1536$\times$2048 RGB images from six different viewpoints and a 3D point cloud computed from the depth images. We followed the data splits in~\cite{4dor}, \ie, 6 videos (4,024 scenes) were used for training and 2 videos (1,332 scenes) for validation, while 2 sequences (1,378 scenes) were kept for testing. The annotation includes 12 subject/object categories in the operating room and 14 interaction categories. For evaluation, we used precision, recall, and  F1 metrics, following the state-of-the-art methods \cite{4dor,LABRADOR,sformer}. 
We evaluate the proposed framework on the public benchmark 4D-OR~\cite{4dor}.
The dataset consists of 10 knee replacement surgery videos, involving 6,734 frames captured by RGB-D cameras.
Each frame includes RGB images with a size of 1536$\times$2048 from six different viewpoints and 3D point clouds computed from the depth map. 
We follow the same data partitioning as in~\cite{4dor}, where 6 videos (4,024 scenes) are for training, 2 videos (1,332 scenes) for validation, and 2 videos (1,378 scenes) for online testing.
The annotations contain 12 subject/object classes and 14 relation classes in ORs.
For evaluation, we use precision, recall, and F1 metrics in line with previous methods~\cite{4dor,LABRADOR}.

% For a more comprehensive evaluation, we also adopted metrics which are widely-used for the Open Images dataset including recall under the IoU threshold of 0.5 (R@50), weighted mean average precision of relationship detection (wmAPrel), and phrase detection (wmAPphr).

\begin{table*}[t!]
\centering
% \footnotesize
\renewcommand{\arraystretch}{1.5}
\renewcommand{\tabcolsep}{0.4mm}
\caption{Detailed comparisons with existing OR-SGG models on 4D-OR test set.}
\label{main_table}
  \smallskip\noindent
  \resizebox{\textwidth}{!}{
\begin{tabular}{c|l|c|ccccccccccccccc}
\hline
  Methods                   & Params  & Metrics  & Assist & Cement & Clean & CloseTo & Cut & Drill & Hammer & Hold & LyingOn & Operate & Prepare & Saw & Suture & Touch & Avg \\ 
\hline
\multirow{3}{*}{4D-OR~\cite{4dor}}      & \multirow{3}{*}{84.8M} &  Precision     & 0.42   &  0.78  & 0.53   &  \textbf{0.97}   &   0.49  &  0.87   &   0.71   &   0.55  & \textbf{1.00}    &   0.55   & 0.62   &  0.69   &   0.60    &   0.41   & 0.68    \\ 
                          &                             &   Recall      & \textbf{0.93}   &  0.78   &   0.63  &  0.89   &  0.49   &   \textbf{1.00}   &   0.89    &  \textbf{0.95}    &   0.99   &   \textbf{0.99}    &    \textbf{0.91}  &    0.91  &   \textbf{1.00}  &   0.69   &   0.87    \\ 
                         &                             &     F1 & 0.58   &  0.78    &    0.57  &   0.93   &  0.49   &  0.93   &   0.79   &   0.70   &   0.99    &   0.71    &   0.74   &  
 0.79   &    0.75    &  0.51     & 0.75    \\ 
 \hline \multirow{3}{*}{LABRAD-OR~\cite{LABRADOR}}  & \multirow{3}{*}{85.8M}   &   Precision     & 0.60   &   0.96   &  0.86    &   0.96   & \textbf{0.91}    &    \textbf{1.00}   &    0.93    &  0.71    &  \textbf{1.00}      &    0.85    &    0.77     &   0.78  &   \textbf{1.00}   &   0.71  & 0.87    \\ 
&         &    Recall      & 0.86   &   0.93     &   0.72    &   \textbf{0.94}   &   0.68  &  0.94   &  \textbf{0.95}   &  \textbf{0.95}    &   \textbf{1.00}    &   \textbf{0.99}    &  \textbf{0.91}    &  0.93   &     \textbf{1.00}   &    \textbf{0.72}   & \textbf{0.90}    \\ 
  &               &   F1 & 0.71   &   0.94     &   0.78    &    \textbf{0.95}   & 0.78    &   \textbf{0.97}  &  0.94      &   0.81   & \textbf{1.00} &   0.91    &   0.84   &  0.85   &     \textbf{1.00}   &   0.71   & 0.88    \\ 
\hline  \multirow{3}{*}{\textbf{\ourmodel}} &  \multirow{3}{*}{\textbf{67.1M}}    &  Precision     & \textbf{0.74}   &   \textbf{1.00}    &  \textbf{0.92}  &   \textbf{0.97}    &   0.86  &   0.98    &   \textbf{0.96}    &  \textbf{0.84}   &   \textbf{1.00}    &   \textbf{0.93}    &    \textbf{0.86}   &   \textbf{0.95}  &  0.96    &  \textbf{0.82}     & \textbf{0.91}    \\ 
  &          &    Recall      & 0.79   & \textbf{0.95}     &   \textbf{0.88}    &  \textbf{0.94}    &  \textbf{0.85}   &     0.94  &    \textbf{0.95}    &   0.82   &    \textbf{1.00}    &      0.90   &  0.88       &  \textbf{0.99}   &      0.95  &   \textbf{0.72}    & \textbf{0.90}    \\
 &           &     F1  & \textbf{0.76}   &  \textbf{0.97}      &  \textbf{0.89}     &   \textbf{0.95}    &  \textbf{0.85}   &   0.96    &  \textbf{0.95}  &   \textbf{0.83}   &   \textbf{1.00}      &   \textbf{0.92}      &    \textbf{0.87}   &  \textbf{0.97}   &   0.95   &  \textbf{0.77}   & \textbf{0.90}    \\ \hline
\end{tabular}}
% \vspace{-20pt}
\end{table*}

% Qualitative results of the 4D-OR model \cite{4dor}, LABRAD-OR \cite{LABRADOR}, S$^2$Former-OR~\cite{sformer}, and our \ourmodel~on the 4D-OR validation set. The circles represent the human/object attribute. Erroneous predicted relationships are shown in red.
% \vspace{-5pt}
\subsection{Implementation Details}
% We trained the proposed model for 80 epochs on 2 RTX3090 GPUs with an AdamW optimizer. The batch size per GPU was set to 2, and the initial learning rate was 5e-5 with a weight decay of 0.0001. We adopted ResNet-50 \cite{resnet} pre-trained on ImageNet~\cite{imagenet} and the DETR encoder \cite{detr} pre-trained on MS-COCO \cite{cocodataset} as the multi-view image encoder. We took two typical viewpoints, view\#1 and view\#6, as model inputs. We achieved the 2D bounding box annotations following \cite{sformer}. During training, the short edges of the input multi-view images were randomly resized to a value between 480 and 800 with the aspect ratio fixed, while they were fixed to 800 during validation and inference. Additionally, we applied color jitter and random horizontal flip to RGB images and shuffled the point cloud data for augmentation. Restricted by the spatial size of the feature maps, we employed 4 different scale kernels in each convolutional kernel group in ViewTemp with kernel size $\{1,3,5,7\}$ for view\#1 and $\{3,5,7,9\}$ for view\#6, respectively. The cross-attention operation In Eq.~\ref{Eq:dynamic} consists of 2 multi-head transformer encoder layers with 4 heads.

%the short edges of the input multi-view images are randomly resized to a value between 480 and 800 with the fixed aspect ratio, while they are fixed to 800 during validation and inference. Additionally, 
We train the proposed model for 60 epochs with an AdamW optimizer. 
The batch size is 4, and the initial learning rate is 5e-5 with a weight decay of 0.0001. The loss weight $\lambda_{c}$ and $\lambda_{t}$ are set to 1 and 0.1, respectively. We adopt ResNet-50 \cite{resnet} pre-trained on ImageNet~\cite{imagenet} as the image encoder.
In our experiments, two typical viewpoints (view\#1 and view\#6) are selected as 2D image inputs to our model, and the number of consecutive frames $l$ is 3. During training, we apply color jitter, random resize, and random horizontal flip to augment 2D images. 
According to the feature map size, we employ kernel size combination $\{1,3,5,7\}$ for view\#1, and $\{3,5,7,9\}$ for view\#6 in ViewTemp.

\begin{figure*}[t!]
	\centering
	\includegraphics[width=.98\linewidth]{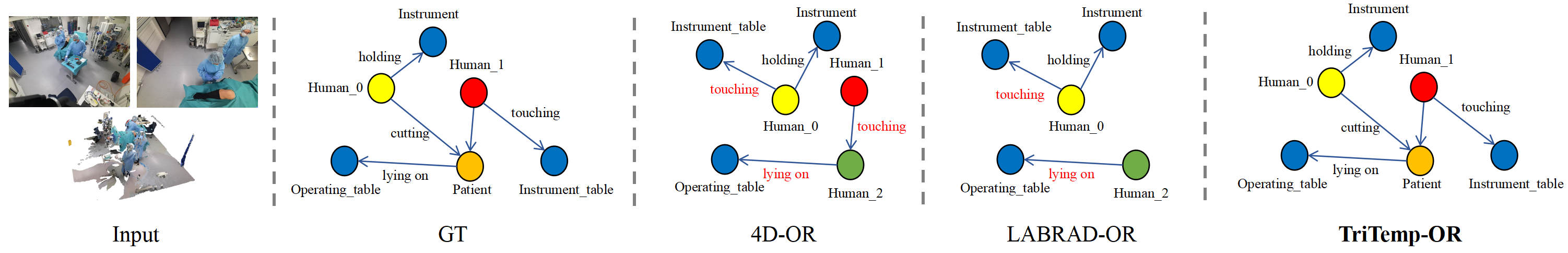}
        % \vspace{-10pt}
	\caption{Qualitative results of our \ourmodel~and existing OR-SGG methods on the 4D-OR validation set. Erroneous predicted relations are shown in red.}
	\label{vis}
    % \vspace{-10pt}
\end{figure*}

% \vspace{-10pt}
\subsection{Comparison with State-of-the-art Methods}

% In Tab. \ref{main_table}, we compared the experimental results of our model with three state-of-the-art (SOTA) baselines for the SGG task in OR, involving 4D-OR \cite{4dor}, LABRAD-OR \cite{LABRADOR}, and S$^2$Former-OR \cite{sformer}. Among them, 4D-OR is a multi-stage model based on VoxelPose \cite{voxelpose} and Group-Free \cite{groupfree}, which utilizes 3D point cloud information and multi-view RGB 2D images for SGG. Based on 4D-OR, LABRAD-OR further integrates temporal information with a memory bank. S$^2$Former-OR and \ourmodel~perform the task in a single stage. Besides, our model benefits from the proposed temporal fusion and knowledge transfer from LLaVA-Med. As it can be observed from the table, \ourmodel~demonstrates promising performance, particularly in average precision and  F1 score, achieving 0.91 and 0.90, respectively. According to the table, our model alleviates the class imbalance problem discussed in the introduction part. Even compared to our closest competitor, S$^2$Former-OR, our model shows a remarkable advantage in recognizing less-frequent surgical actions in the dataset such as `Touching' and `Assisting', by 14\% and 5\%, respectively. We attribute this to the introduction of the transferred knowledge from LLaVA-Med. 
\tabref{main_table} compares the experimental results of our method with two existing OR-SGG models, 4D-OR~\cite{4dor} and LABRAD-OR~\cite{LABRADOR}.
Both of them use multi-view images together with point cloud samples as input.
Furthermore, LABRAD-OR and our~\ourmodel~also utilize temporal information to enhance the predictions in consecutive frames.
As observed in~\tabref{main_table}, \ourmodel~shows promising performance with only two views as input, especially in the average precision and  F1 score of 0.91 and 0.90, respectively.
\figref{vis} exhibits the scene graphs generated by existing OR-SGG methods on the 4D-OR validation set.
Benefiting from the transferred knowledge of LLaVA-Med and temporal interaction, our model also mitigates the class-imbalance problem.
Even compared to LABRAD-OR, our model demonstrates a remarkable advantage in recognizing less-frequent relations such as `Saw' and `Clean', with improvements of 12\% and 11\%.

% has added temporal fusion and knowledge transfer from LLaVA-Med with only an extra 10.7\% parameters. Although there is trivial improvement in common relational classes such as `Lying on' and `Close to', our model predicts more accurately in some rare and challenging surgical actions, including `Touching' and `Assisting'. 
% Specifically, \ourmodel~improves the  F1 by 14\% and 5\%, respectively, alleviating the class imbalance problem on the 4D-OR dataset. 
% As shown in Fig. \ref{vis},  our model can predict the rare interaction classes more robustly compared to state-of-the-art SGG-OR models.

\vspace{15pt}
\noindent % 确保从页面边缘开始布局
\begin{minipage}{.5\textwidth}
    \centering
    \includegraphics[scale=.19]{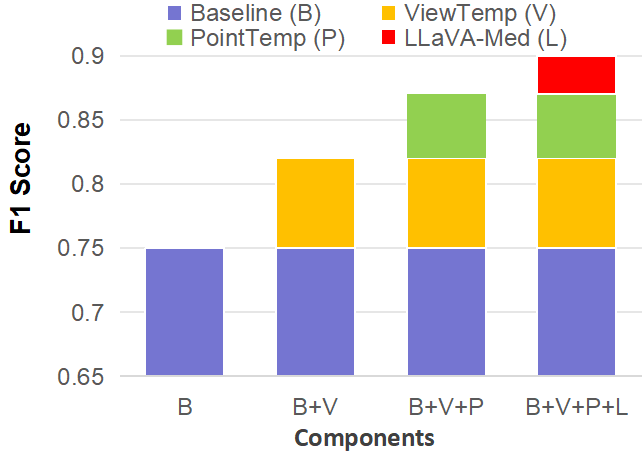}
    \captionof{figure}{Ablation study for the contribution of each component.}
    \label{ablation2}
\end{minipage}%
\hfill % 在两个minipage之间添加空白，使它们分布在左右两侧
\begin{minipage}{.48\textwidth}
    \centering
    \footnotesize
    \renewcommand{\arraystretch}{.9}
    \setlength\tabcolsep{2pt}
    \begin{tabular}{l|c|c|c}
    \hline
     View\#1/View\#6 & P $\uparrow$ & R $\uparrow$ & F1 $\uparrow$ \\ \hline
     $\left \{1, 3, 5, 7 \right \}/ \left \{1, 3, 5, 7 \right \}$ & 0.89 & 0.88 & 0.88 \\
     $\left \{1, 3, 5, 7 \right \}/ \left \{3, 5, 7, 9 \right \}$ & \textbf{0.91} & \textbf{0.90} & \textbf{0.90} \\
     $\left \{1, 3, 5, 7 \right \}/ \left \{5, 7, 9, 9 \right \}$ & 0.90 & 0.87 & 0.88 \\
     $\left \{3, 5, 7, 9 \right \}/ \left \{3, 5, 7, 9 \right \}$ & 0.88 & 0.86 & 0.87\\
    $\left \{5, 7, 9, 9 \right \}/ \left \{3, 5, 7, 9 \right \}$ & 0.85 & 0.82 & 0.83 \\
     $\left \{3, 3, 3, 3 \right \}/ \left \{3, 3, 3, 3 \right \}$ & 0.87 & 0.84 & 0.85\\ 
     \hline
    \end{tabular}
    \captionof{table}{Ablations on kernel size combinations in ViewTemp. P and R represent Precision and Recall.}
    \label{ablation1}
    % \vspace{10pt}
\end{minipage}

\subsection{Ablation Study}

\noindent\textbf{Ablation Study for Each Component.} We study the contribution of each component to the model performance in Fig. \ref{ablation2}. Results show that all components, including ViewTemp, PointTemp, as well as the knowledge transfer from LLaVA-Med, contribute to our final result. Among these components, ViewTemp has the most remarkable impact, bringing in an increase of $7\%$ in average F1 score. This shows that multi-view temporal information is pivotal for the spatial modeling of the OR environment. PointTemp comes next, reflecting the importance of dynamic information provided by temporal 3D point cloud features in surgical action recognition. The LLaVA-Med text embedding also improves the prediction, which proves that the transferred knowledge from medical LLMs enhances the comprehension of intraoperative actions.

\begin{figure}[b!]
    \centering
    \vspace{-10pt}
    \begin{adjustbox}{left=110mm}
    % 第一张图
    \begin{minipage}{.22\textwidth}
        \centering
        \includegraphics[width=\linewidth]{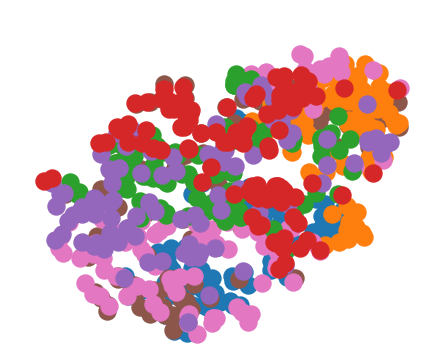}
    \end{minipage}
    % 第二张图
    \begin{minipage}{.22\textwidth}
        \centering
        \includegraphics[width=\linewidth]{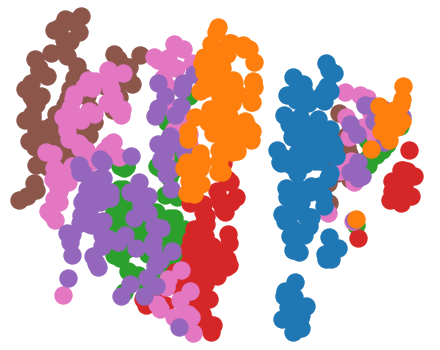}
    \end{minipage}
    \hspace{10pt}
    \begin{minipage}{.3\textwidth}
        \centering
        \footnotesize
        \renewcommand{\arraystretch}{.9}
        \setlength\tabcolsep{1pt}
    \begin{tabular}{l|c|c|c}
        \hline
         Embeddings & P $\uparrow$ & R $\uparrow$ & F1 $\uparrow$ \\ \hline
         No Embeddings & 0.88 & 0.86 & 0.87 \\
         LLaVA-Med~\cite{llavamed} & \textbf{0.91} & \textbf{0.90} & \textbf{0.90} \\
         CLIP~\cite{clip} & 0.87 & 0.86 & 0.86 \\
         \hline
    \end{tabular}
    \end{minipage}
     \end{adjustbox}
    \caption{\textbf{Left:} t-SNE of CLIP latent space. \textbf{Middle:} t-SNE of LLaVA-Med latent space. \textbf{Right:} Ablations on the impact of the knowledge transfer from different pre-trained LLMs. P and R represent Precision and Recall.}
    \label{last}
\end{figure}

\noindent\textbf{t-SNE of CLIP and LLaVA-Med.} \figref{last} shows the 3D t-SNE results of different sentences in the latent space from CLIP~\cite{clip} and LLaVA-Med~\cite{llavamed}. The points represent the text embeddings of the prompted sentences. Sentences with the same intraoperative actions are denoted with the same color. As illustrated in the figure, points of different colors from CLIP embeddings intermingle indiscriminately. In contrast, within the LLaVA-Med embedding space, points sharing the same color coalesce into clusters, demonstrating that LLaVA-Med has better discrimination for different relations in surgery compared to CLIP.

\noindent\textbf{Knowledge Transfer from LLMs.} 
% We utilize LLaVA-Med in this paper, based on its large-scale pre-training on bio-medical data. We report the performance of \ourmodel~with and without text embeddings given by different LLMs in Tab. \ref{ablation3}. Experimental results show that the knowledge transfer from LLaVA-Med, the biomedical LLM, outperforms all the others. It is noticeable that the text embeddings from CLIP do not bring an improvement to the model without text embedding, which could be attributed to the knowledge gap between open-vocabulary LLMs and surgical understanding tasks.
In our experiments, we use LLaVA-Med~\cite{llavamed} pre-trained on large-scale biomedical data to perform knowledge transfer to enhance the context awareness of our model and alleviate the relation class imbalance.
To validate the effectiveness of pre-trained LLMs to the OR-SGG task, we display the results of~\ourmodel~with and without text embeddings from different LLMs in~\figref{last}.
Notably, the text embeddings from the text encoder of CLIP~\cite{clip} bring no benefit to our model, which may be caused by the prior knowledge gap between the open vocabulary and the biomedical semantics.

\noindent\textbf{Combination of Different Kernel Sizes in ViewTemp.} 
\tabref{ablation1} compares the effect of kernel size combinations for different viewpoints in ViewTemp.
When increasing the kernel sizes from $\{1,3,5,7\}$ to $\{3,5,7,9\}$, view\#6 has a positive impact on the accuracy, while view\#1 does the opposite.
It is attributed to the broader perspective in view\#1 while view\#6 provides a more focused viewpoint.
Our model achieves the optimal results with the combination of $\{1,3,5,7\}$ and $\{3,5,7,9\}$.
Besides, using the scale-adaptive partition obtains better results than the single-scale kernel sizes $\{3,3,3,3\}$ across different perspectives.

% \vspace{-5pt}
\section{Conclusion}
% \vspace{-3pt}
This paper presents \ourmodel, an end-to-end tri-modal framework for the OR-SGG task. In one fold, we introduce ViewTemp and PointTemp to capture the temporal dynamics of multi-view images and point clouds. 
The bi-modal features are subsequently integrated with relation-aware feature unification to predict subject-object relations.
On the other fold, we distill the semantic knowledge from the biomedical LLM, obtaining the tri-modal representation and further alleviating the class-imbalance problem. 
% Experimental results show that our method consistently outperforms the state-of-the-art methods, endorsing a potential application for effective surgical workflow monitoring.
Experimental results illustrate that our method consistently outperforms the state-of-the-art methods, endorsing a potential application for improving surgical procedure efficiency in ORs.

    \small
    \bibliographystyle{splncs04}
    \bibliography{ref}

\end{document}